\documentclass[10pt,twocolumn,letterpaper]{article}

\usepackage{cvpr}              
\definecolor{cvprblue}{rgb}{0.21,0.49,0.74}
\usepackage[pagebackref,breaklinks,colorlinks,allcolors=cvprblue]{hyperref}

\usepackage[accsupp]{axessibility}  

\usepackage{booktabs}
\usepackage{rotating}
\usepackage{multirow} 
\usepackage{makecell} 
\usepackage{colortbl} 
\usepackage{arydshln} 
\usepackage[dvipsnames]{xcolor}
\usepackage{tabularx}
\usepackage{pifont} 
\usepackage{amssymb}
\usepackage{subcaption} 
\usepackage{times}
\usepackage{soul}
\usepackage[utf8]{inputenc}
\usepackage{caption}
\usepackage{graphicx}
\usepackage{amsmath}
\usepackage{amsthm}
\usepackage{booktabs}
\usepackage[switch]{lineno}
\usepackage{makecell}
\usepackage{xspace}
\usepackage{wrapfig}
\usepackage{tcolorbox}

\newcommand{\xmark}{\ding{55}}%
\newcommand{\noo}{\textcolor{red}{\xmark}}
\newcommand{\yes}{\textcolor{OliveGreen}{\checkmark}}

\colorlet{colorFst}{Green!25}       
\colorlet{colorSnd}{SpringGreen!45} 
\colorlet{colorTrd}{Yellow!30}      
\colorlet{colorLow}{darkgray!30}    

\newcommand{\bff}{\boldsymbol{f}}
\newcommand{\bfy}{\boldsymbol{y}}

\colorlet{colorYes}{SkyBlue!30}
\colorlet{colorNo}{Lavender!30}
\colorlet{colorMaybe}{YellowOrange!30}

\newcommand{\cyes}{\cellcolor{colorYes}}
\newcommand{\cno}{\cellcolor{colorNo}}
\newcommand{\cmaybe}{\cellcolor{colorMaybe}}

\definecolor{promptcolor}{HTML}{D1D0F2}
\definecolor{promptcolorheader}{HTML}{bdbcec}
\newcommand{\promptbox}[2]{
    \begin{tcolorbox}[
        top=0.3em,bottom=0.3em,left=0.5em,right=0.5em,
        toptitle=0.3em,bottomtitle=0.2em,boxsep=0pt,
        colframe=promptcolorheader,colback=promptcolor!50,boxrule=0.5pt,
        title={\footnotesize \fontfamily{zi4}\selectfont #1}
    ]
        \footnotesize
        {\fontfamily{zi4}\selectfont #2}
    \end{tcolorbox}
}

\newcommand{\ourmethod}{UniMatch\xspace}

\def\paperID{} 
\def\confName{CVPR}
\def\confYear{2026}

\title{Universal 3D Shape Matching via Coarse-to-Fine Language Guidance}

\author{Qinfeng Xiao$^1$ \quad Guofeng Mei$^2$ \quad Bo Yang$^1$\footnotemark[1] \quad Liying Zhang$^1$ \quad Jian Zhang$^3$ \quad Kit-lun Yick$^{1}$\footnotemark[1]\\
$^1$Hong Kong Polytechnic University, HK SAR \\ $^2$Fondazione Bruno Kessler, Italy \\ $^3$University of Technology Sydney, Australia\\
{\tt\small qin-feng.xiao@connect.polyu.hk, \{bo.yang,kit-lun.yick\}@polyu.edu.hk}
}

\begin{document}


\twocolumn[{%
  \renewcommand\twocolumn[1][]{#1}%
  \maketitle
  \vspace{-5mm}
  \begin{center}
    \includegraphics[width=1.0\linewidth]{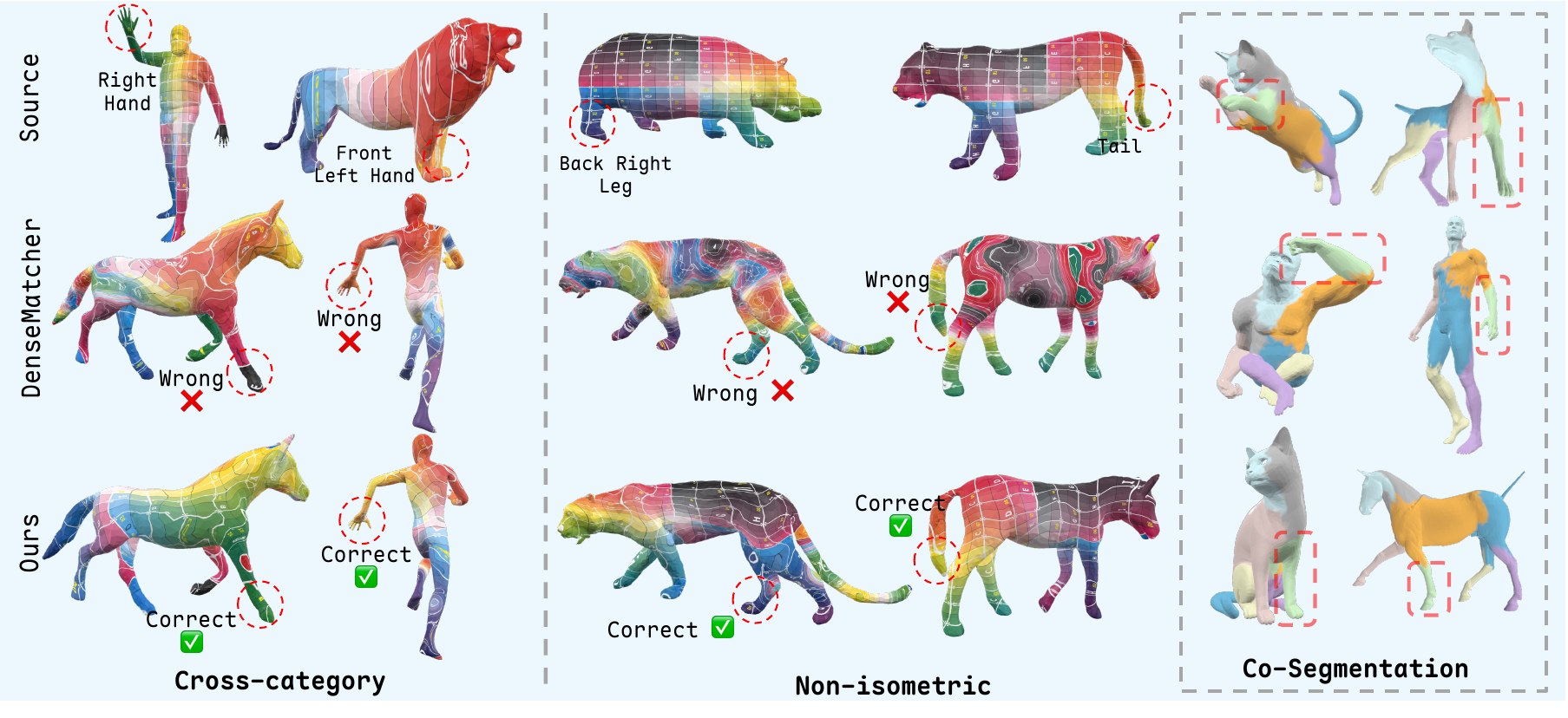}
    \captionof{figure}{{We propose \ourmethod, a semantic-aware, coarse-to-fine framework for 3D shape matching.} Our method yields high-quality semantic correspondences under challenging scenarios, \eg, cross-category shapes and non-isometric deformations. Besides,  \ourmethod learns semantically consistent cross-shape features, enabling efficient co-segmentation for distinct categories.}
    \label{fig:teaser}
    \vspace{-2mm}
  \end{center}
}]

\renewcommand{\thefootnote}{\fnsymbol{footnote}}
\footnotetext[1]{Corresponding authors}

\begin{abstract}
Establishing dense correspondences between shapes is a crucial task in computer vision and graphics, while prior approaches depend on near-isometric assumptions and homogeneous subject types (\ie, only operate for human shapes). However, building semantic correspondences for cross-category objects remains challenging and has received relatively little attention. To achieve this, we propose \textbf{\ourmethod}, a semantic-aware, coarse-to-fine framework for constructing dense semantic correspondences between strongly non-isometric shapes without restricting object categories. The key insight is to lift ``coarse'' semantic cues into ``fine'' correspondence, which is achieved through two stages. In the ``coarse'' stage, we perform class-agnostic 3D segmentation to obtain non-overlapping semantic parts and prompt multimodal large language models (MLLMs) to identify part names. Then, we employ pretrained vision language models (VLMs) to extract text embeddings, enabling the construction of matched semantic parts. In the  ``fine'' stage, we leverage these coarse correspondences to guide the learning of dense correspondences through a dedicated rank-based contrastive scheme. Thanks to class-agnostic segmentation, language guiding, and rank-based contrastive learning, our method is versatile for universal object categories and requires no predefined part proposals, enabling universal matching for inter-class and non-isometric shapes. Extensive experiments demonstrate \ourmethod consistently outperforms competing methods in various challenging scenarios.
\end{abstract}
    
\section{Introduction}
\label{sec:intro}

Shape matching aims to establish dense correspondences between 3D shapes and underpins a wide range of applications, such as texture transfer~\citep{ezuz2017deblurring}, parametric human modeling~\citep{loper2023smpl}, robotic manipulation~\citep{zhu2025densematcher}, and shape interpolation~\citep{eisenberger2021neuromorph,cao2024spectral}. A long-standing and effective paradigm represents point-to-point maps in a compact spectral representation through functional maps~\citep{ovsjanikov2012functional}, enabling elegant regularization and efficient optimization. Deep functional map approaches further process geometric descriptors and operators in an end-to-end fashion~\citep{litany2017deep,donati2020deep,li2022learning,sun2023spatially,cao2023unsupervised}. Despite strong performance, they typically rely on a near-isometric assumption, and thus degrade under substantial non-isometric deformations or topological noises~\citep{dyke2020shrec,lahner2016shrec,ovsjanikov2012functional}. Moreover, purely geometric cues struggle to support inter-class matching, where correspondences between inter-class variations must bridge semantic relations~\citep{dutt2024diffusion,zhu2025densematcher}.

In parallel, visual foundation models (VFMs) have unlocked rich, transferable semantics, and surprisingly robust correspondence cues from 2D images~\citep{oquab2023dinov2,tang2023emergent}. Building on these priors, recent methods lift or distill 2D features into 3D for shape matching: Diff3F~\citep{dutt2024diffusion} decorates untextured meshes with diffusion models and learn correspondences via intermediate diffusion features; DenseMatcher~\citep{zhu2025densematcher} learns 3D semantic correspondences through SD-DINO features~\citep{zhang2023tale} and manual annotated parts for robot manipulation; and a zero-shot pipeline, ZSC, estimates dense correspondences by feeding coarse part-level correspondences into a spectral-based refinement algorithm~\citep{abdelreheem2023zero}. Although compelling, these methods remain limited: they are not versatile for in-the-wild objects in fully unsupervised settings, while ZSC requires predefined part proposals, and DenseMatcher requires manual part annotations.

To address the limitations, we propose \textbf{\ourmethod}, a semantically aware, coarse-to-fine framework for computing semantic correspondences without restricting object categories. \ourmethod operates in two stages. At the ``coarse'' stage, we first employ an effective class-agnostic 3D part segmentation algorithm to obtain non-overlapping semantic parts without predefined part proposals or category hints. Then we employ the multimodal reasoning capabilities of GPT-5~\citep{openai_gpt5_systemcard} to identify the name of each part and build coarse part-level correspondence by computing language embeddings using FG-CLIP~\citep{xie2025fg}. At the ``fine'' stage, we extend the functional map framework by leveraging semantic feature fields calculated using SD-DINO~\citep{zhang2023tale} and the coarse correspondences. Specifically, we propose a novel group-wise rank-based contrastive loss that effectively enforces semantic consistency across shapes without requiring explicit positive/negative hints, unlike conventional contrastive losses. A comparison with major baselines on key features of our proposed method is summarized in \cref{tab:compare}.

Our contributions are as follows:
\begin{itemize}
    \item We present \textbf{\ourmethod}, a semantic-aware, \textit{coarse-to-fine} shape matching framework for universal object categories without predefined part priors.
    \item We build coarse part-level correspondences via class-agnostic 3D part segmentation, MLLM prompting, and FG-CLIP language embeddings.
    \item We compute point-to-point correspondences by extending the conventional functional map framework with \textit{semantic feature fields} and our proposed \textit{group-wise rank-based contrastive loss} that enforces semantic consistency without explicit positives/negatives.
    \item Extensive experiments show consistent gains of our method on various challenging scenarios, including inter-/non-/near-isometric settings.
\end{itemize}

\begin{table}[ht]
    \centering
    \footnotesize
    \caption{{Overview of our method and competing methods.}}
    \vspace{-2mm}
    \label{tab:compare}
    \begin{tabular}{lccc}
\toprule
Method & \makecell{Semantic\\Features} & \makecell{Language\\Guided} & Inter-class \\ 
\midrule
URSSM~\citep{cao2023unsupervised} \textcolor{gray}{[ToG 2023]} & \cno{No~\noo} & \cno{No~\noo} & \cno{No~\noo} \\ 
Diff3F~\citep{dutt2024diffusion} \textcolor{gray}{[CVPR 2024]} & \cyes{Yes~\yes} & \cno{No~\noo} & \cmaybe{Worse} \\
ZSC~\citep{abdelreheem2023zero} \textcolor{gray}{[SIGGR. Asia 2023]} & \cno{No~\noo} & \cno{No~\noo} & \cyes{Yes~\yes} \\
DenseMatcher~\citep{zhu2025densematcher} \textcolor{gray}{[ICLR 2025]} & \cyes{Yes~\yes} & \cno{No~\noo} & \cyes{Yes~\yes} \\
\ourmethod (ours) & \cyes{Yes~\yes} & \cyes{Yes~\yes} & \cyes{Yes~\yes} \\
\bottomrule
\end{tabular}
\end{table}

\section{Related Work}
\label{sec:related}

\subsection{Shape Correspondences}


Establishing 3D shape correspondences is crucial for various applications, and most methods can be categorized into two dominant paradigms: the functional map approach and its variants, and recent advances that leverage large-scale models to explore semantic and part-guided correspondences. Functional map~\citep{ovsjanikov2012functional} methods model correspondences as compact linear operators in spectral domains, applying geometric constraints for regularization. Recent advances use learned descriptors and dedicated objectives to improve matching accuracy~\citep{litany2017deep,donati2020deep,li2022learning,sun2023spatially,cao2023unsupervised,roufosse2019unsupervised,halimi2019unsupervised,melzi2019zoomout}. Though effective for near-isometric deformations, they depend on mesh assumptions and weakly capture high-level semantic structure, resulting in degraded performance under strong non-isometric or cross-category variations. 



\noindent\textbf{Semantic correspondences.} Due to the inefficiency of conventional approaches for tackling non-isometric deformations, recent works have explored alternative approaches to find \textit{semantic correspondences} between shapes, which are expected to capture high-level semantic relations beyond handcrafted geometric descriptors. Meanwhile, diffusion models~\citep{rombach2022high} and VFMs have demonstrated strong capabilities in encoding semantic-rich features, enabling the computation of semantic correspondences across shapes. Since these models are designed for 2D images, existing methods leverage multi-view rendering and aggregation to obtain 3D semantic features. Diff3F~\citep{dutt2024diffusion} utilizes a pre-trained diffusion model and DINO~\citep{oquab2023dinov2} to extract multi-view features, which are then aggregated to form 3D features for correspondence estimation. DenseMatcher~\citep{zhu2025densematcher} employs SD-DINO~\citep{zhang2023tale} features extracted from rendered images and learns matching with the help of manual part annotations.

\noindent\textbf{Part-guided correspondences.} 3D part segmentation can be categorized into two main approaches: text-prompted segmentation and class-agnostic segmentation. Text-prompted part segmentation lift 2D visual detectors, \eg BLIP~\citep{li2022blip}, from multi-view renderings and fuse the segmentation results~\citep{liu2023partslip,zhou2023partslip++,abdelreheem2023satr,kim2024partstad}. Instead of identifying certain parts using text prompts, class-agnostic methods decompose the entire object into semantic parts by distilling SAM masks~\citep{liu2025partfield,huang2024openins3d,deng2025geosam2,yang2024sampart3d} or training in an end-to-end fashion~\citep{ma2025p3,zhu2025partsam}. To achieve zero-shot and inter-class shape matching, ZSC~\citep{abdelreheem2023zero} proposes using part segmentation to build part-level relations and refine them to produce dense correspondences. However, it relies on predefined part proposals and text prompts for each category, limiting its generalization to in-the-wild objects. DenseMatcher~\citep{zhu2025densematcher} proposes to use a contrastive loss to enforce consistency between annotated parts. Although it incorporates visual features and semantic consistency, it requires manually labeled semantic parts, which limits its broader applications in open-world scenarios.

\subsection{Large-Scale Models}

Recent advances in large-scale models have significantly pushed the boundaries of multimodal understanding~\citep{yin2024survey,li2025survey}. Multimodal Large Language Models (MLLMs) such as GPT-4~\citep{achiam2023gpt}, GPT-5~\citep{openai_gpt5_systemcard}, LLaVA~\citep{liu2023visual}, and QWen-VL~\citep{bai2023qwen} have demonstrated remarkable abilities in comprehending and reasoning about visual data in a conversational context. Concurrently, Vision-Language Models (VLMs) like CLIP~\citep{radford2021learning} and its fine-grained variant FG-CLIP~\citep{xie2025fg}, as well as SigLIP~\citep{tschannen2025siglip}, have excelled at learning powerful, generalizable visual representations from vast amounts of image-text data. These models provide robust semantic features that can be transferred to various downstream tasks. We employ the powerful multimodal reasoning capabilities of GPT-5 to identify semantic part names and leverage FG-CLIP to extract fine-grained language embeddings for building coarse part-level correspondences. 

\section{Method}
\label{sec:method}

\begin{figure*}[ht]
    \centering
    \vspace{-1mm}
    \includegraphics[width=1.0\linewidth]{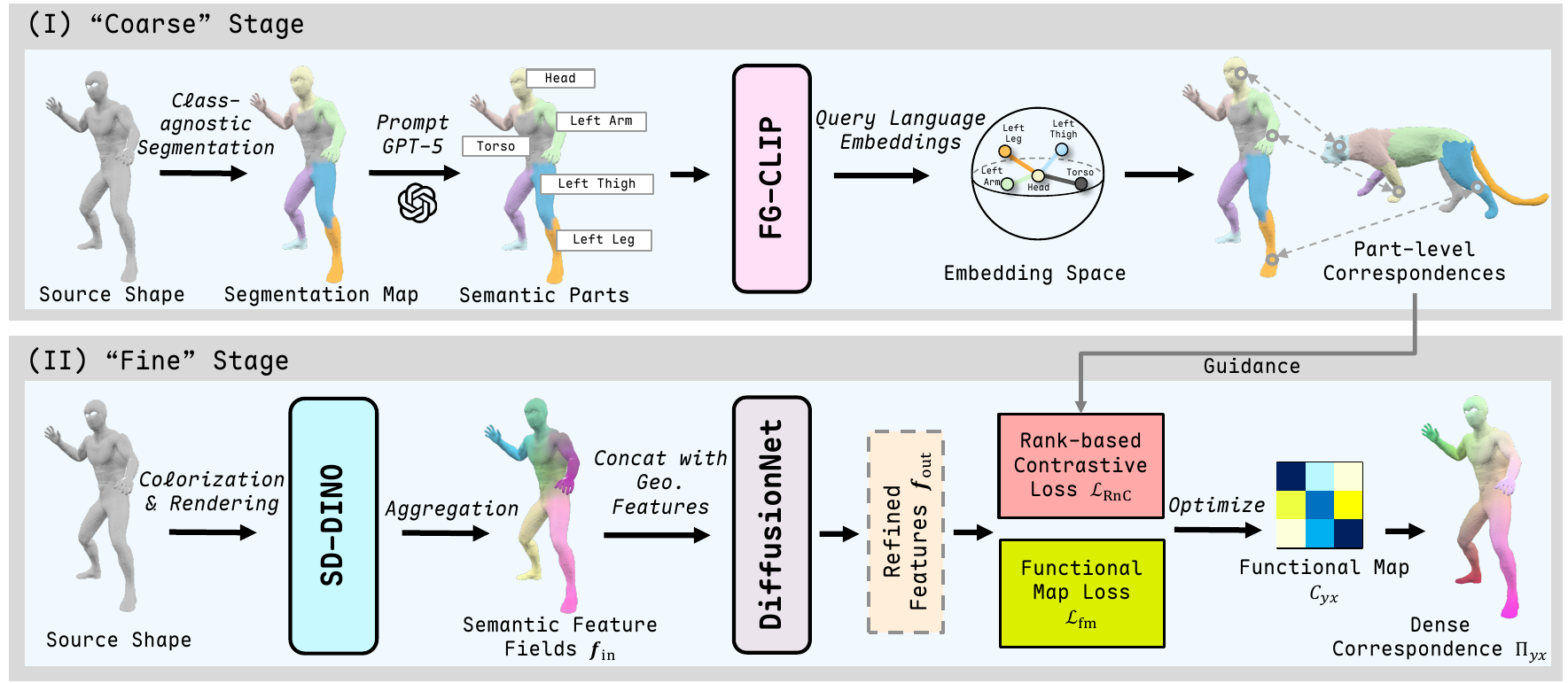}
    \vspace{-3mm}
    \caption{Our framework consists of two stages: (i) \textbf{Coarse stage}: class-agnostic 3D segmentation (PartField) produces non-overlapping semantic regions. Multi-view rendering and multimodal large language model (MLLM) prompting (using GPT-5 and FG-CLIP) assign part names and map them to unified language embeddings, enabling implicit, robust coarse correspondences across shapes. (ii) \textbf{Fine stage}: the coarse semantic matches guide dense correspondence learning within an extended functional map pipeline, leveraging SD-DINO semantic feature fields and a novel group-wise rank-based contrastive loss to enforce semantic consistency. \ourmethod operates without predefined part priors and generalizes across object categories and non-isometric deformations.
    }
    \label{fig:framework}
    \vspace{-4mm}
\end{figure*}

Our method, as illustrated in \cref{fig:framework}, is a two-stage framework that operates in a coarse-to-fine manner. In the first, \textit{coarse} stage, we prompt MLLMs to retrieve semantic regions. We then utilize language embeddings to establish semantic-rich coarse correspondences between shape pairs. In the second, \textit{fine} stage, these coarse correspondences ``supervise" a functional map framework. This process is optimized using a dedicated group-wise ranking-based contrastive loss to yield the final dense correspondences.

\subsection{The ``coarse'' stage: semantic region relations}

\indent\textbf{Class-agnostic part segmentation.} Many works utilize SAM~\citep{kirillov2023segment} to achieve text-prompted 3D part segmentation~\citep{liu2023partslip,abdelreheem2023satr,abdelreheem2023zero,ma2024find}, \ie using 2D visual grounding models and SAM to obtain 2D masks and aggregate them into 3D masks. In this paper, we consider an effective class-agnostic 3D segmentation method, PartField~\citep{liu2025partfield}, instead of text-prompted segmentation. We obtain the corresponding non-intersecting 3D masks for each, representing a semantic region for the input shape $\mathcal X$. The reasons are threefold: \emph{(i) the segmentation accuracy of text-referred methods can be disastrous for existing shape matching benchmarks, as they are often untextured and low resolution; (ii) text-prompted methods require explicit names of semantic parts, which are limited for open-vocabulary objects; (iii) even given part definitions, they cannot cover the whole shape, leading to incomplete matches for the downstream shape-matching task; (iv) PartField demonstrates on-the-fly inference speed due to the feedforward 3D architecture, while text-prompted 3D segmentation algorithms require complex rendering, grounding and aggregation processes.} Given an input shape $\mathcal X$ and the number of parts $n_\mathcal{R}$, we denote the segmentation result as $\mathcal R_x$.


\noindent\textbf{Multi-modal semantic region prompting.} Despite the efficiency and high accuracy of class-agnostic segmentation of PartField, the semantic regions obtained have no corresponding part names. To tackle this, we prompt MLLMs to obtain a part name for each semantic region, thereby establishing coarse correspondences from the input shape pair. In detail, we first render the input shape with 3D masks into multi-view images and corresponding 2D masks using a differentiable renderer, by sequencing the cameras counterclockwise around the shape. Then, we overlay each 2D mask on the original image and prompt GPT-5~\citep{openai_gpt5_systemcard} to obtain the name of the shaded region. We discard too small masks, \ie, less than $5\%$ pixels out of the whole object. Finally, we aggregate part names into the 3D domain using known camera parameters to obtain the final semantic region proposal $\mathcal R_x$. More details can be found in \cref{suppl:mllm}. Compared to obtaining semantic regions with class-dependent part proposals~\citep{abdelreheem2023zero}, our method is more flexible in handling open-world 3D shapes and requires no predefined part proposals. Additionally, ZSC requires prompt MLLM during inference time for testing shapes. In contrast, our method only prompts MLLM for training.


\noindent\textbf{Solving ambiguity by language.} We attempt to build coarse correspondences among semantic regions. However, building explicit correspondences like ZSC~\citep{abdelreheem2023zero} is challenging and lacks flexibility because the input shape pair may not share the same semantic partitions, especially for cross-category shape matching. Besides, part names produced by MLLMs can be ambiguous, \eg ``mouth'' of a human and ``muzzle'' of a dog should be semantically matched while they have totally different language outputs. To keep semantic consistency and robustness towards MLLM's outputs, we build implicit correspondences between semantic regions with language embeddings. In detail, we map part names into a unified embedding space by feeding them into FG-CLIP~\citep{xie2025fg} to obtain language embeddings $\mathcal E\in \mathbb R^{C_\text{lang}}$ where $C_\text{lang}$ is the embedding dimension. Later, the similarity between the two parts can be measured using a common distance measurement. Compared with ZSC, which requires heavy MLLM prompting during inference to obtain part relations, our method only prompts GPT-5 for training data curation. We show that smooth, continuous language embeddings benefit the optimization process rather than explicit and hard-coded correspondences and reveal the semantic-rich ranks among parts in \cref{sec:rank_contrast}.

\subsection{Fine Stage}

\noindent\textbf{Functional map pipeline.} The functional map~\citep{ovsjanikov2012functional} pipeline models 3D shape correspondences as linear operators between functional spaces, typically using a spectral basis. Instead of point-to-point mapping, it computes a compact representation that can be converted back when needed~\citep{melzi2019zoomout}. We adopt the advanced variant of the functional map framework, URSSM~\citep{cao2023unsupervised}, in this work. Specifically, given precomputed per-vertex features $\boldsymbol f_\text{in}$ for $n$ vertices, a trainable refiner, \eg, DiffusionNet~\citep{sharp2022diffusionnet}, $\mathcal F$ produces refined features: $\boldsymbol f_\text{out} = \mathcal F_\theta(\boldsymbol f_\text{in})$, where $\mathcal F_\theta(\cdot):\mathbb R^{n \times C_\text{in}} \rightarrow \mathbb R^{n \times C_\text{out}}$. The refiner $\mathcal F$ and the functional map $C_{yx}$ are jointly optimized with: (1) data preserving loss $\mathcal L_\text{data}$ to preserve input features $\bff_\text{out}$; (2) regularization loss $\mathcal L_\text{reg}$ to ensure mathematical properties like bijectivity and orthogonality; (3) coupling loss $L_\text{couple}$ to ensure the consistency between soft correspondences (calculated by the cosine similarity of $\boldsymbol f_\text{out}$) and the functional map $C_{yx}$. The final functional map objective is:
\begin{equation}
\label{eq:fmaps}
    \mathcal L_\text{fm} = \mathcal L_\text{data} + \lambda_\text{reg} \cdot \mathcal L_\text{reg} + \lambda_\text{couple} \cdot \mathcal L_\text{couple}.
\end{equation}
This pipeline, while effective for near-isometric shapes, fails in non-isometric shape matching and heterogeneous subjects~\citep{ovsjanikov2012functional,abdelreheem2023zero,zhu2025densematcher}. To this end, we enhance it by incorporating \emph{semantic feature fields} and a \emph{group-wise rank-based contrastive loss} for robust matching across categories and non-isometric deformations.

\paragraph{Semantic feature fields.}
Previous functional map methods~\citep{halimi2019unsupervised,roufosse2019unsupervised,cao2023unsupervised} typically utilize geometric descriptors, \eg wave kernel signature (WKS)~\citep{aubry2011wave} to feed the refiner $\mathcal F$. Inspired by the success of VFMs~\citep{radford2021learning,caron2021emerging,oquab2023dinov2,rombach2022high}, recent work attempts to lift 2D semantic knowledge to 3D~\citep{dutt2024diffusion,abdelreheem2023zero,zhu2025densematcher}. Similar to DenseMatcher~\citep{zhu2025densematcher}, we build our semantic feature fields using SD-DINO~\citep{zhang2023tale} but discard the use of positional encoding, since it results in disastrous performance. Besides, we perform view-consistent texture synthesis for uncolored shapes using SyncMVD~\citep{liu2024text}, unlike textured meshes used in DenseMatcher. Then we render $K$ multi-view RGB images with uniformly distributed elevation and azimuthal angles in $[0^\circ,360^\circ)$. To extract semantic features, we use SD-DINO to get low-resolution features and employ the FeatUp upscaler~\citep{fu2024featup} to upsample them. Lastly, given known camera parameters, we back-project and aggregate 2D semantic features into the 3D domain from all visible views. Finally, we input concatenation of the geometric descriptors $\boldsymbol f_\text{geo}$ and semantic features $\boldsymbol f_\text{sem}$ to the refiner $\mathcal F$:
\begin{equation}
    \boldsymbol f_\text{in} = \texttt{Concat}(\boldsymbol f_\text{geo}, \boldsymbol f_\text{sem}).
\end{equation}

\begin{figure}[ht]
    \centering
    \includegraphics[width=0.95\linewidth]{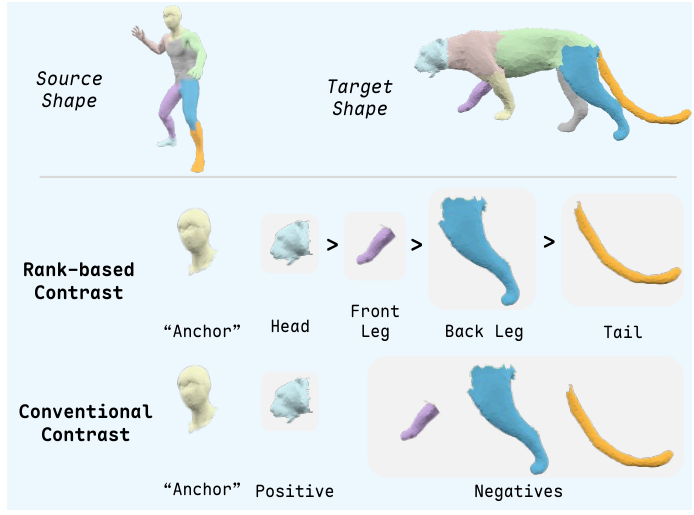}
    \vspace{-3mm}
    \caption{\emph{Intuition of rank-based contrastive loss.} Compared with standard contrastive loss, rank-based contrastive loss effectively utilizes the ordinal hints to optimize features.}
    \label{fig:part_order}
    \vspace{-4mm}
\end{figure}

\noindent\textbf{Rank-based contrastive loss.}~\label{sec:rank_contrast}
We design a rank-based contrastive loss for learning semantic-aware correspondences.
Given the shape pair $\mathcal X$ and $\mathcal Y$ with per-vertex features $\boldsymbol{f}_x$ and $\boldsymbol{f}_y$, we use the implicit coarse correspondences to ``supervise'' the optimization for dense correspondences. The most straightforward solution is to adopt contrastive learning, \eg, SupCon loss~\citep{khosla2020supervised}, widely used in representation learning. However, conventional contrastive losses require explicit hints to distinguish positive and negative samples, which are not available in our case. Besides, this scheme fails to recognize the underlying continuous relations between semantic regions provided by language embeddings. To tackle this, we utilize Rank-n-Contrastive (RnC) loss~\citep{zha2023rank} to learn continuous representations that ordered semantic relations between parts (see \cref{fig:part_order}).

\vspace{1mm}
\noindent\emph{Preliminaries of RnC loss.}
To align the distances in the feature space ordered by similarities provided by label hints (\ie, language embeddings in our case), RnC loss first ranks all samples based on label distances regarding an anchor sample, and then contrasts them according to their relative rankings. Specifically, given an anchor $\bff^x_i$, RnC loss constructs the likelihood of a reference $\bff^y_j$ conditioned on a set of negative samples $\mathcal S_{i,j} := \{\bff^y_k|\mathrm d(\bfy_i,\bfy_k)\geq\mathrm d(\bfy_i,\bfy_j)\}$, which contains all samples whose ranks (defined by the label distance measure $\mathrm d(\cdot, \cdot)$) to the anchor are lower than that of the reference. The likelihood is defined as:
\begin{equation}
    \mathbb P(\bff^y_j|\bff^x_i,\mathcal S_{i,j}) = \frac{\exp(\text{sim}(\bff^x_i,\bff^y_j)/\tau)}{\sum_{\bff^y_k\in \mathcal S_{i,j}}\exp(\text{sim}(\bff^x_i,\bff^y_k)/\tau)},
\end{equation}
where $\text{sim}(\cdot, \cdot)$ is the similarity measure and $\tau$ is the temperature parameter. Intuitively, maximizing $\mathbb P(\bff^y_j|\bff^x_i,\mathcal S_{i,j})$ increases the likelihood that samples of higher ranks than the reference $\bff^y_j$ and the reference are closer to the anchor $\bff^x_i$ than all negative samples in $\mathcal S_{i,j}$. 

\vspace{1mm}
\noindent\emph{The proposed group-wise RnC loss.} 
To supervise dense correspondences with the RnC loss, one can contrast per-vertex features using coarse semantic rankings derived from language embeddings; 
however, this poses $O(n_x\times n_y)$ time and memory complexity, and assumes vertex independence, ignoring the inherent grouping into semantic regions that provides structured ordinal relations. We propose a \textit{group-wise Rank-n-Contrastive loss} that overcomes these limitations by contrasting at the semantic group level rather than per-sample, reducing complexity to $O(n_x\times n_R)$, where $n_R$ is the number of regions ($n_R \ll n_y$), while explicitly modeling inter-group dependencies through embedding-based distances to enforce continuous semantic consistency. 

\begin{figure}
    \centering
    \includegraphics[width=0.98\linewidth]{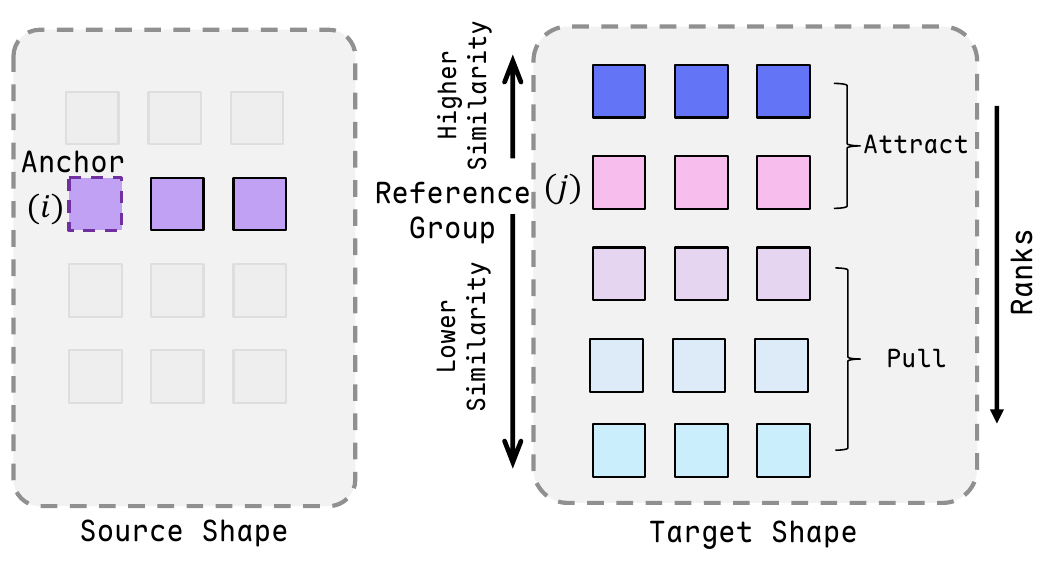}
    \vspace{-3mm}
    \caption{\textbf{Illustration of our group-wise Rank-n-Contrastive Loss.} For each \textcolor[RGB]{193, 161, 243}{anchor} feature, we traverse the semantic regions as \textcolor[RGB]{247, 189, 236}{reference} features, and group the negative samples by their semantic similarities (lower than reference). The darker the color, the closer the semantic similarity.}
    \label{fig:rank}
    \vspace{-3mm}
\end{figure}

Specially, given an anchor feature $\boldsymbol f^x_i$ from the source shape and a reference group, denoted as $\mathcal G_j^y$, from the target shape, negatives $\mathcal S_{i,j}:=\{\boldsymbol{f}^y_k|k\neq i,\mathrm d(\mathcal{E}_i,\mathcal{E}_k) \geq \mathrm d(\mathcal{E}_i,\mathcal{E}_j)\}$ are dynamically grouped by language embedding distances $\mathrm d(\cdot,\cdot)$, yielding per-group likelihoods:
\begin{equation}
    \mathbb P(\mathcal G_j^y|\bff^x_i,\mathcal S_{i,j}) = \frac{\sum_l\exp(\text{sim}(\bff^x_i,\bff^y_l)/\tau)}{\sum_{\bff^y_k\in \mathcal S_{i,j}}\exp(\text{sim}(\bff^x_i,\bff^y_k)/\tau)},
\end{equation}
where $\tau$ is the temperature parameter, and the numerator is the summation of similarities of the anchor and the reference group $\mathcal G_j^y$. The group-wise loss aggregates negative log-likelihoods over target regions:
\begin{align}
    \label{eq:rnc_loss}
    \ell_\text{RnC}^{(i)}(\mathcal X, \mathcal Y) = \frac{1}{n_\mathcal{R}}\sum_{j=1}^{n_\mathcal{R}} -\log\mathbb P(\bff^y_j|\bff^x_i,\mathcal S_{i,j}).
\end{align}
Finally, the group-wise RnC loss is the average over source anchors, calculated by:
\begin{equation}
    \mathcal L_\text{RnC} = \frac{1}{n_x}\sum_{i=1}^{n_x} \ell_\text{RnC}^{(i)}(\mathcal X, \mathcal Y).
\end{equation}
This enables scalable, annotation-free alignment, preserves RnC’s ranking under grouped, language-guided supervision, and yields smoother, more robust correspondences than vertex-wise/independent contrasts (see \cref{tab:ablation}, \cref{fig:rank}).


\section{Experiments}
\label{sec:exp}

\begin{figure*}[ht]
    \centering
    \includegraphics[width=0.9\linewidth]{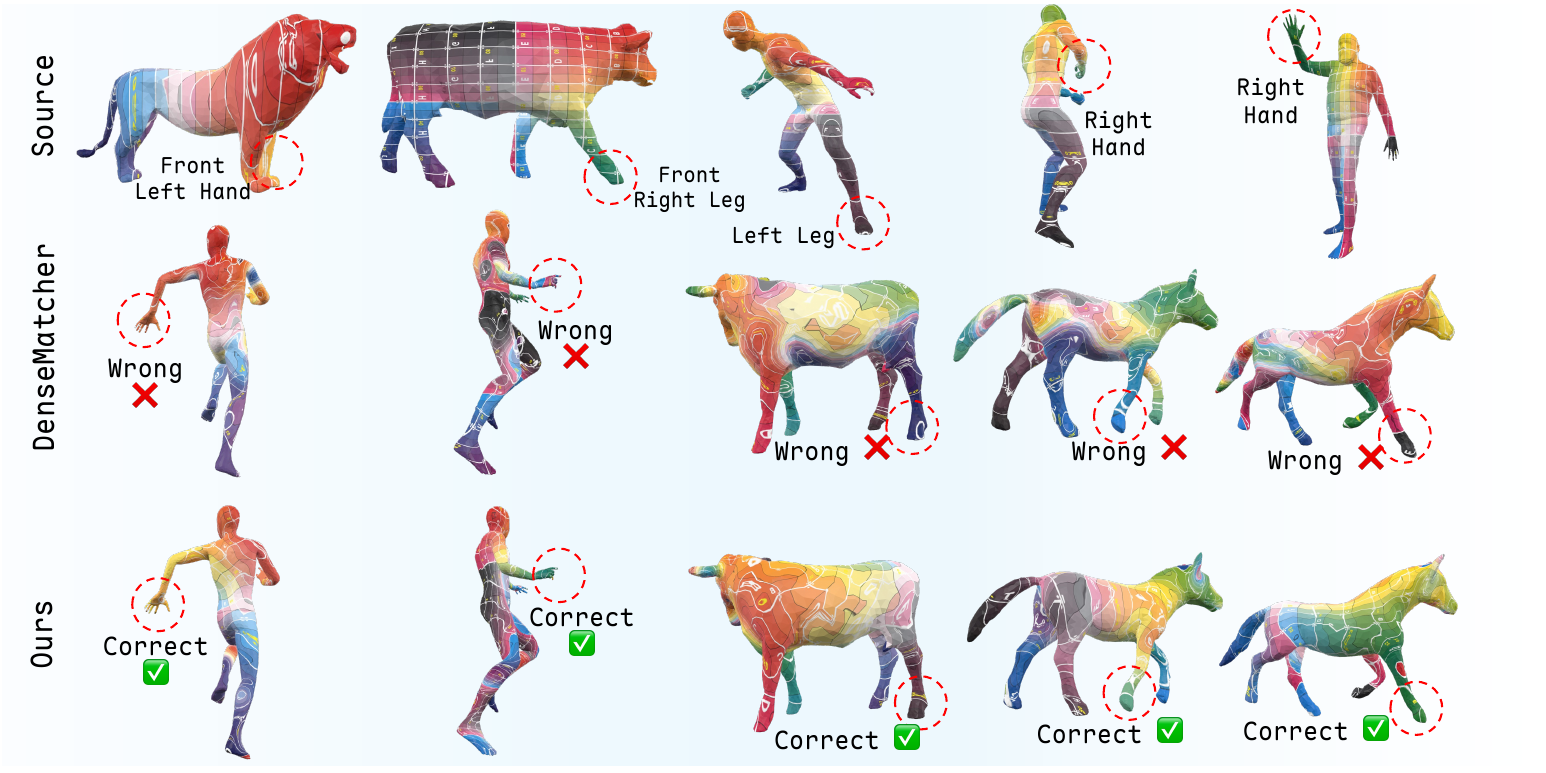}
    \vspace{-3mm}
    \caption{Visual comparison with the state-of-the-art method, DenseMatcher~\citep{zhu2025densematcher} on inter-class shape matching. \ourmethod demonstrates semantic consistency and smooth correspondences for challenging cross-category matching.}
    \label{fig:exp_inter_class}
    \vspace{-3mm}
\end{figure*}

\subsection{Inter-class shape matching}
\noindent\textbf{Setup.} We evaluate our method and baselines on three challenging cross-category benchmarks, Strongly Non-Isometric Shapes (SNIS)~\citep{abdelreheem2023zero}, TOSCA~\citep{bronstein2008numerical}, and SHREC07~\citep{Giorgi2007SHapeRC}. 
Details of datasets and baselines can be found in \cref{suppl:dataset_interclass} and \cref{suppl:baseline}, respectively. Since no valid dense annotations are provided for inter-class matching, we follow prior works~\citep{abdelreheem2023zero,dutt2024diffusion} to compute the average geodesic error on sparse annotated vertices.

\begin{table}[ht]
    \centering
    \tabcolsep 9pt
    \small
    \caption{Experimental results of inter-class shape matching. The best and the second-best are shown in \textcolor{SkyBlue}{blue} and \textcolor{YellowOrange}{orange} respectively.}
    \vspace{-2mm}
    \label{tab:exp_inter_class}
    \begin{tabular}{lccc}
\toprule
Method & SNIS & TOSCA & SHREC07 \\
\midrule
\multicolumn{4}{l}{\cellcolor[HTML]{EEEEEE}{\textit{Axiomatic Methods}}} \\
ZoomOut~\citep{melzi2019zoomout}  &   0.51  & 0.55 & 0.57 \\ 
\midrule
\multicolumn{4}{l}{\cellcolor[HTML]{EEEEEE}{\textit{Functional Map Methods}}} \\
URSSM~\citep{cao2023unsupervised}  & 0.49      & 0.53 & 0.49 \\ 
SimpFMap~\citep{magnet2024memory}   & 0.51      & 0.53 & 0.56 \\ 
\midrule
\multicolumn{4}{l}{\cellcolor[HTML]{EEEEEE}{\textit{Semantic Methods}}} \\
Diff3F~\citep{dutt2024diffusion}  & 0.57    &  0.45 & 0.50 \\
ZSC~\citep{abdelreheem2023zero}   & 0.36 & 0.56 & 0.60 \\
DenseMatcher~\citep{zhu2025densematcher}  & \cmaybe{0.28} & \cmaybe{0.30} & \cmaybe{0.39} \\
\ourmethod (ours) & \cyes{\textbf{0.19}}   & \cyes{\textbf{0.23}} & \cyes{\textbf{0.37}} \\
\bottomrule
\end{tabular}
\end{table}

\noindent\textbf{Results.} \cref{tab:exp_inter_class} reports the numerical results of \ourmethod and representative baselines on three cross-category benchmarks. \ourmethod achieves the lowest average geodesic error on all three cross-category benchmarks: $0.19$ (SNIS), $0.23$ (TOSCA), and $0.37$ (SHREC07).
This performance substantially outperforms all previous methods, including DenseMatcher, ZSC, and other functional map-based or axiomatic approaches. 
In detail, DenseMatcher, although competitive, falls short ($0.28$–$0.39$) compared to \ourmethod, suggesting that \ourmethod's universal part segmentation and group-wise contrastive loss contribute critically to more substantial semantic alignment. The visual comparison between \ourmethod and DenseMatcher is shown in \cref{fig:exp_inter_class}. Functional map approaches (\eg, URSSM, SimpFMap) exhibit higher errors ($0.49$–$0.56$), validating that geometric-only solutions lack sufficient flexibility for inter-class matching. This consistent margin affirms the robustness and effectiveness of the semantic-aware matching paradigm in highly challenging cross-category scenarios. The superior performance paves the way for universal, scalable matching technology for in-the-wild 3D objects.

\begin{figure}[ht]
    \centering
    \includegraphics[width=0.95\linewidth]{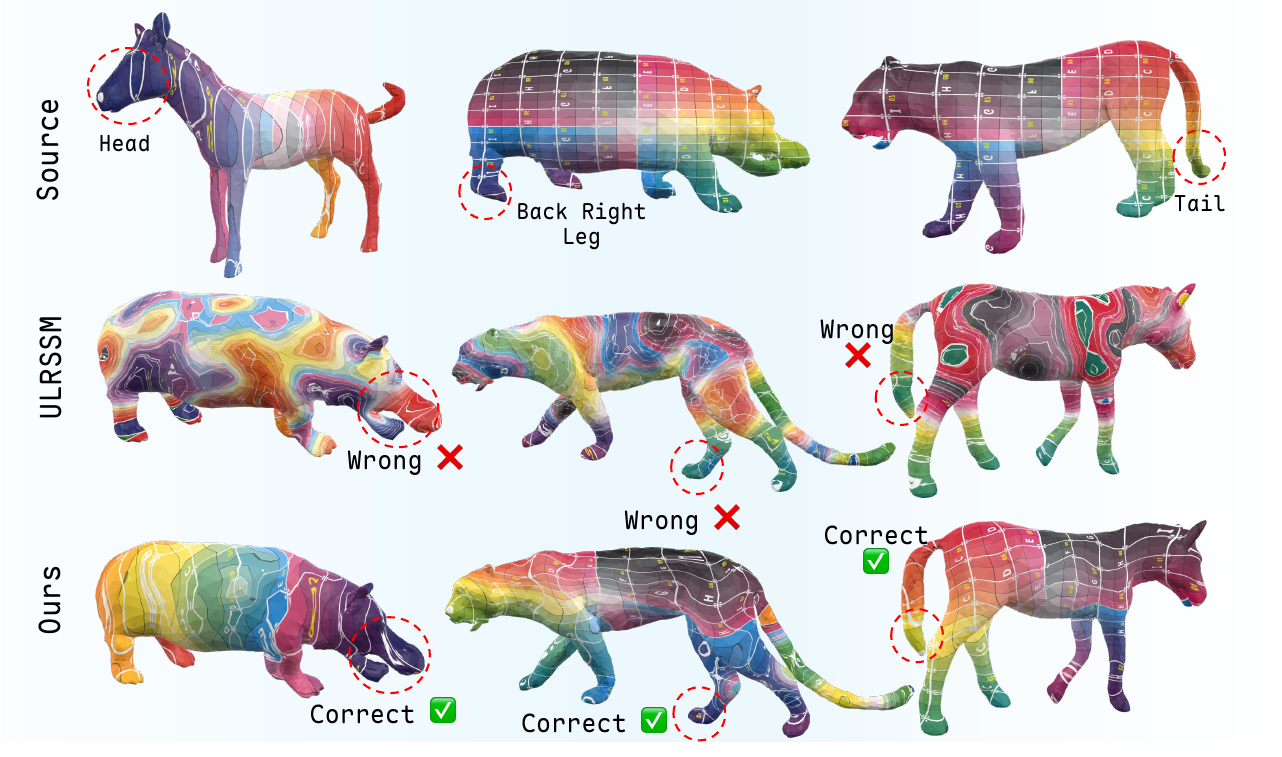}
    \vspace{-5mm}
    \caption{{Visual comparison with the representative functional map method, URSSM~\citep{cao2023unsupervised} on non-isometric shapes.}}
    \label{fig:vis_smal}
    \vspace{-3mm}
\end{figure}

\subsection{Non-Isometric Shape Matching} 
\noindent\textbf{Setup.} We evaluate \ourmethod's performance on non-isometric shape matching using two datasets, SMAL~\citep{zuffi20173d} and TOPKIDS~\citep{lahner2016shrec}, which present substantial shape variations and deformations (\eg, quadruped animals and synthetic human children in different poses). We use average geodesic error ($\times100$) as the evaluation metric. More details of datasets and baseline implementation can be found in \cref{suppl:dataset_noniso} and \cref{suppl:baseline}, respectively.

\begin{table}[ht]
    \centering
    \small
    \tabcolsep 14pt
    \caption{Experimental results of non-isometric shape matching.}
    \vspace{-2mm}
    \label{tab:exp_non_isometric}
    \begin{tabular}{lcc}
\toprule
Method & SMAL & TOPKIDS \\ 
\midrule
\multicolumn{3}{l}{\cellcolor[HTML]{EEEEEE}{\textit{Axiomatic Methods}}} \\
ZoomOut~\citep{melzi2019zoomout}  & 38.4 & 33.7 \\ 
Smooth Shells~\citep{eisenberger2020smooth}  & 36.1 & 11.8 \\ 
DiscreteOp~\citep{ren2021discrete}  & 38.1 & 35.5 \\ 
\midrule
\multicolumn{3}{l}{\cellcolor[HTML]{EEEEEE}{\textit{Functional Map Methods}}} \\
UnsupFMNet~\citep{halimi2019unsupervised}  & -    & 38.5 \\
SURFMNet~\citep{roufosse2019unsupervised}  & -    & 48.6 \\
AttentiveFMaps~\citep{li2022learning}  & 5.4 & 23.4 \\
URSSM~\citep{cao2023unsupervised}  & 6.0  & 8.9 \\ 
\midrule
\multicolumn{3}{l}{\cellcolor[HTML]{EEEEEE}{\textit{Semantic Methods}}} \\
Diff3F~\citep{dutt2024diffusion}  & 28.4 & 31.0 \\
DenseMatcher~\citep{zhu2025densematcher}  & \cyes{\textbf{4.7}}  & \cmaybe{6.2} \\
\ourmethod (ours) &  \cmaybe{4.8} & \cyes{\textbf{5.9}} \\
\bottomrule
\end{tabular}
\vspace{-4mm}
\end{table}

\noindent\textbf{Results.} \cref{tab:exp_non_isometric} reports average geodesic errors ($\times 100$) on SMAL and TOPKIDS. \ourmethod achieves average geodesic errors of $4.8$ on SMAL and $5.9$ on TOPKIDS, which is comparable or slightly better than DenseMatcher ($4.7$ and $6.2$), and notably better than most other baselines. Functional map-based methods such as URSSM ($6.0$ and $8.9$), AttentiveFMaps ($5.4$ and $23.4$), and SURFMNet, as well as semantic approaches like Diff3F ($28.4$ and $31.0$), show higher errors. The visualization results compared with the representative functional map method, URSSM, is illustrated in \cref{fig:vis_smal}. Axiomatic methods, such as ZoomOut and Smooth Shells ($38.4$/$33.7$ and $36.1$/$11.8$), are significantly less effective in this benchmark. The superior performance indicates that \ourmethod is highly robust to substantial non-isometric deformations. The results confirm that our approach effectively overcomes limitations of prior functional map-based and semantic methods, especially for shapes that deviate strongly from near-isometric assumptions.

\subsection{Near-Isometric Shape Matching}

\indent\textbf{Setup.} We use three widely-used benchmarks (FAUST~\citep{bogo2014faust}, SCAPE~\citep{anguelov2005scape}, SHREC19~\citep{melzi2019shrec}) for near-isometric shape matching. Following~\cite{cao2023unsupervised}, we consider the remeshed versions from~\cite{donati2020deep} and adopt average geodesic error ($\times100$) as the evaluation metric. 
Details of datasets and baselines can be found in \cref{suppl:dataset_iso} and \cref{suppl:baseline}, respectively.

\begin{table}[ht]
\tabcolsep 8pt
    \centering
    \small
    \caption{Experimental results of near-isometric shape matching.}
    \vspace{-2mm}
    \label{tab:exp_isometric}
    \begin{tabular}{lccc}
\toprule
Method & FAUST & SCAPE & SHREC19 \\
\midrule
\multicolumn{4}{l}{\cellcolor[HTML]{EEEEEE}{\textit{Axiomatic Methods}}} \\
BCICP~\citep{ren2018continuous}  & 6.4 & 11.0 & 8.0 \\ 
ZoomOut~\citep{melzi2019zoomout}  & 6.1 &  7.5 & 7.8 \\ 
Smooth Shells~\citep{eisenberger2020smooth}  & \cmaybe{2.5} &  4.7 & 7.6 \\ 
\midrule
\multicolumn{4}{l}{\cellcolor[HTML]{EEEEEE}{\textit{Functional Map Methods}}} \\
UnsupFMNet~\citep{halimi2019unsupervised}   & 4.8 &  9.6 & 11.1 \\
SURFMNet~\citep{roufosse2019unsupervised}   & \cmaybe{2.5} &  6.0 & 4.8 \\
URSSM~\citep{cao2023unsupervised}  & \cyes{1.6} &  \cyes{1.9} & 5.7 \\ 
\midrule
\multicolumn{4}{l}{\cellcolor[HTML]{EEEEEE}{\textit{Semantic Methods}}} \\
Diff3F~\citep{dutt2024diffusion}  & 20.7 & 22.1 & 26.3 \\
DenseMatcher~\citep{zhu2025densematcher}  & \cyes{1.6}  & \cmaybe{2.0}  & \cyes{\textbf{3.1}} \\
\ourmethod (ours) &  \cyes{\textbf{1.6}}  & \cyes{\textbf{1.9}} & \cmaybe{3.2} \\
\bottomrule
\end{tabular}
\end{table}

\noindent\textbf{Results.} \cref{tab:exp_isometric} shows the average geodesic errors achieved by our proposed method and baselines. Among all methods, \ourmethod achieves state-of-the-art performance, matching or exceeding the best baseline methods on FAUST and SCAPE, and very close to the best on SHREC19. Compared to semantic-driven methods like Diff3F and DenseMatcher, \ourmethod substantially reduces the geodesic error, confirming its greater accuracy. 
The results suggest that \ourmethod is robust to near-isometric deformations, not just cross-category or non-isometric challenges, making it universally applicable for dense shape matching tasks.

\subsection{Ablation Studies}

We ablate our components and design choices on SNIS, TOSCA, and SHREC07.
%

\noindent\textbf{Language embedding calculation.} We conducted an ablation study on language feature extraction to investigate the effects of different language embedding models, \eg CLIP~\citep{radford2021learning}, SigLip~\citep{tschannen2025siglip}, and the method used in this paper, FG-CLIP~\citep{xie2025fg}. We design three variants that utilize language embeddings from the three models, respectively, and keep the remaining components unchanged. As shown in \cref{tab:ablation}, FG-CLIP (our choice) achieves the lowest geodesic errors ($0.19$/$0.23$/$0.37$), matching or slightly outperforming SigLip and clearly better than CLIP, especially on TOSCA. This demonstrates FG-CLIP's superior capacity to deliver fine-grained, continuous semantic signals, vital for robust cross-category correspondence learning.

\noindent\textbf{Semantic feature fields.} To investigate the effectiveness of semantic feature fields, we ablate to evaluate the performance of the functional map with and without semantic feature fields. The first variant utilizes only geometric features, while the second variant employs both geometric and semantic features. As demonstrated in \cref{tab:ablation}, removing semantic features and retaining only geometric features results in significantly higher errors ($0.49$/$0.53$/$0.49$), highlighting the limitation of geometric-only representations for semantic matching. Integrating semantic feature fields, as proposed, drops the errors to $0.22$/$0.26$/$0.39$, confirming their critical role in bridging semantic gaps.

\noindent\textbf{Rank-based contrastive loss.} We also conducted an ablation study on leveraging rank-based contrastive loss to examine the effectiveness for inter-class shape matching. Compared with the variant without RnC loss, the proposed group-wise rank-based contrastive loss further enhances performance, reducing errors to $ 0.19$, $0.23$, and $0.37$, indicating that harnessing the relative ordering and continuous similarity from language embeddings enables more consistent and universal correspondence alignment. We also compare our method with the SupCon loss~\citep{khosla2020supervised}, by taking the top-1 similar sample as the ``pseudo'' positive according to language embeddings. Employing SupCon loss yields average geodesic errors of $0.21$ (SNIS), $0.29$ (TOSCA), and $0.40$ (SHREC07), notably higher than our group-wise RnC loss. This result reveals that SupCon loss, which relies on discrete positive selection, is less effective in capturing the continuous, semantic-rich relations between regions provided by language embeddings.


\begin{table}[ht]
    \centering
    \small
    \caption{{Quantitative results of ablation study.}}
    \vspace{-2mm}
    \label{tab:ablation}
    \begin{tabular}{lccc}
\toprule
 & SNIS & TOSCA & SHREC07 \\ 
\midrule
\multicolumn{4}{l}{\cellcolor[HTML]{EEEEEE}{\textit{Language Embedding}}} \\
CLIP          &    0.21   & 0.26 & 0.37\\ 
SigLip        &    0.19   & 0.24 & 0.37\\ 
FG-CLIP (ours)         &    \textbf{0.19}   & \textbf{0.23} & \textbf{0.37} \\ 
\midrule
\multicolumn{4}{l}{\cellcolor[HTML]{EEEEEE}{\textit{Semantic Feature Fields}}} \\
w/o semantic        &   0.49   & 0.53 & 0.49 \\
w. semantic (ours)    &   \textbf{0.22}   & \textbf{0.26} & \textbf{0.39} \\
\midrule
\multicolumn{4}{l}{\cellcolor[HTML]{EEEEEE}{\textit{Rank-based Contrastive Loss}}} \\
SupCon loss & 0.21 & 0.29 & 0.40 \\
w/o contrastive loss        &  0.22    & 0.26 & 0.39 \\
w. contrastive loss (ours)    &  \textbf{0.19}    & \textbf{0.23} & \textbf{0.37} \\
\bottomrule
\end{tabular}
\vspace{-3mm}
\end{table}

\subsection{Co-segmentation and In-the-wild Objects}

\begin{figure}[ht]
    \centering
    \includegraphics[width=1.0\linewidth]{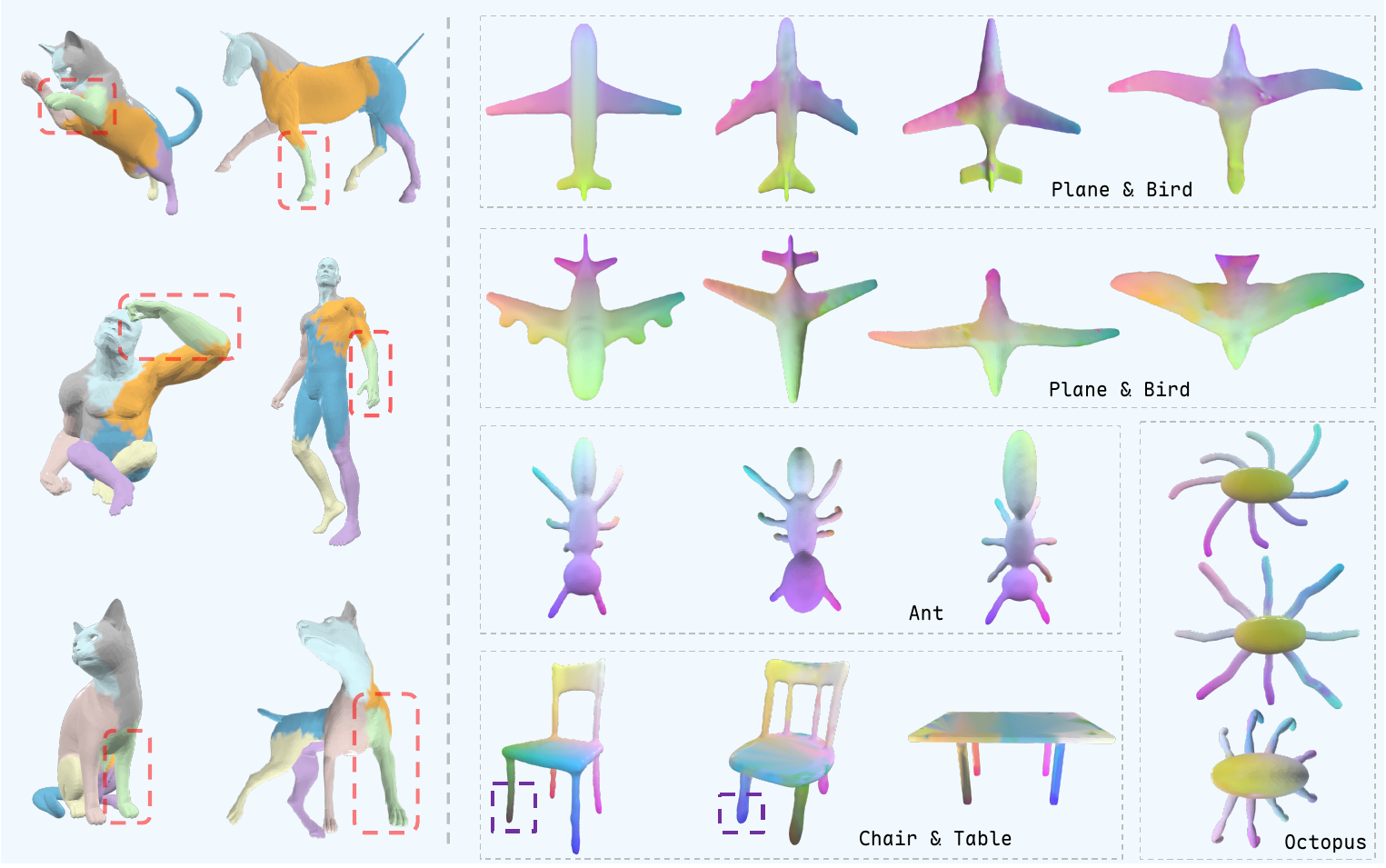}
    \vspace{-5mm}
    \caption{{\textit{Left:} Co-segmentation results across human and animals. \textit{Right:} matching results on in-the-wild objects.}}
    \label{fig:exp_coseg}
    \vspace{-3mm}
\end{figure}

We further explore the semantic consistency of learned features through a co-segmentation task. Specifically, we first adopt agglomerative clustering to segment an anchor shape with vertex connectivity. For the target shape, we initialize K-Means clustering based on the centroids obtained from the clustering result of the anchor shape and perform the K-Means clustering on the features of the target shape. As illustrated in \cref{fig:exp_coseg} (left), although \ourmethod is not delicately designed for segmentation, our method surprisingly emerges semantic consistency across shapes with different topologies and categories. The emerging property demonstrates that \ourmethod is semantic-aware.

We also conducted experiments on a broader range of objects to test the capabilities of matching for in-the-wild objects. Specifically, we evaluate our method on selected categories from SHREC07, including plane, bird, ant, octopus, chair, and table. Other categories have too fuzzy geometric characteristics, \eg, vase and mechanical part, and thus are excluded from this experiment. Despite no consistent correspondence annotations being available for these categories, we illustrate the matching results in \cref{fig:exp_coseg} (right). \ourmethod shows impressive semantic consistency across a wide range of challenging object categories, \eg, wings of a plane and a bird are correctly matched. We also investigate a failure example for the chair category, \ie, the legs are matched in the wrong order (see the purple square). The reason for such an issue is that all legs of the chair can be noted as the word ``leg'' and the algorithm can not infer the correct ``leg'' order from the given inputs. We will fix this issue by considering object orientation in future work.

\section{Conclusion}

We introduce \ourmethod, a semantic-aware, coarse-to-fine 3D shape matching framework that generalizes across object categories and handles strong non-isometric deformations. \ourmethod employs class-agnostic part segmentation, MLLM prompting, and fine-grained language embeddings to establish dense correspondences from coarse guidance without part priors. Experiments show \ourmethod consistently outperforms existing functional map and semantic-driven methods, opening the door to universal semantic shape matching for graphics, robotics, and other domains.

\noindent\textbf{Acknowledgement.} 
We acknowledge financial support from the Research Grant Council (project account PolyU 15606922) and thank the University Research Facility in Big Data Analytics (UBDA) at the Hong Kong Polytechnic University for providing the computing resources that contributed to this research.

{
    \small
    \bibliographystyle{ieeenat_fullname}
    \bibliography{main}
}

\clearpage
\setcounter{page}{1}
\maketitlesupplementary

\section{Implementation Details}
\label{suppl:details}

\subsection{Details of MLLM prompting}
\label{suppl:mllm}

\begin{figure}[ht]
    \centering
    \includegraphics[width=1.0\linewidth]{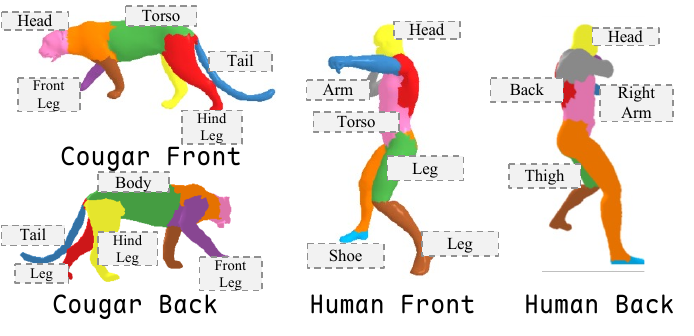}
    \caption{Examples of mask rendering and part names obtained from GPT-5.}
    \label{fig:gpt_prompt}
\end{figure}

After class-agnostic segmentation, we first render 3D objects with 3D masks into 2D images, by a front view and a back view, as shown in \cref{fig:gpt_prompt}. We discard too small masks, \ie, less than $5\%$ pixels out of the whole object. This ratio is selected empirically to avoid misleading GPT-5. Similar to Find3D~\citep{ma2024find}, we later feed the images into GPT-5 and the following prompt to obtain part names:
\promptbox{Infer region names}{
- What is the name of the part that is masked as [COLOR]? If you cannot find the part visible or are not sure, just say unknown. Only output one word or one phrase.
}

Finally, we aggregate part names into the 3D domain using known camera parameters to obtain the final semantic region proposal. As shown in \cref{fig:gpt_prompt}, although the same concept can be expressed in different ways (\eg, body and torso), our method can flexibly match them.

\subsection{Details of \ourmethod}

\begin{figure}[ht]
    \centering
    \includegraphics[width=1.0\linewidth]{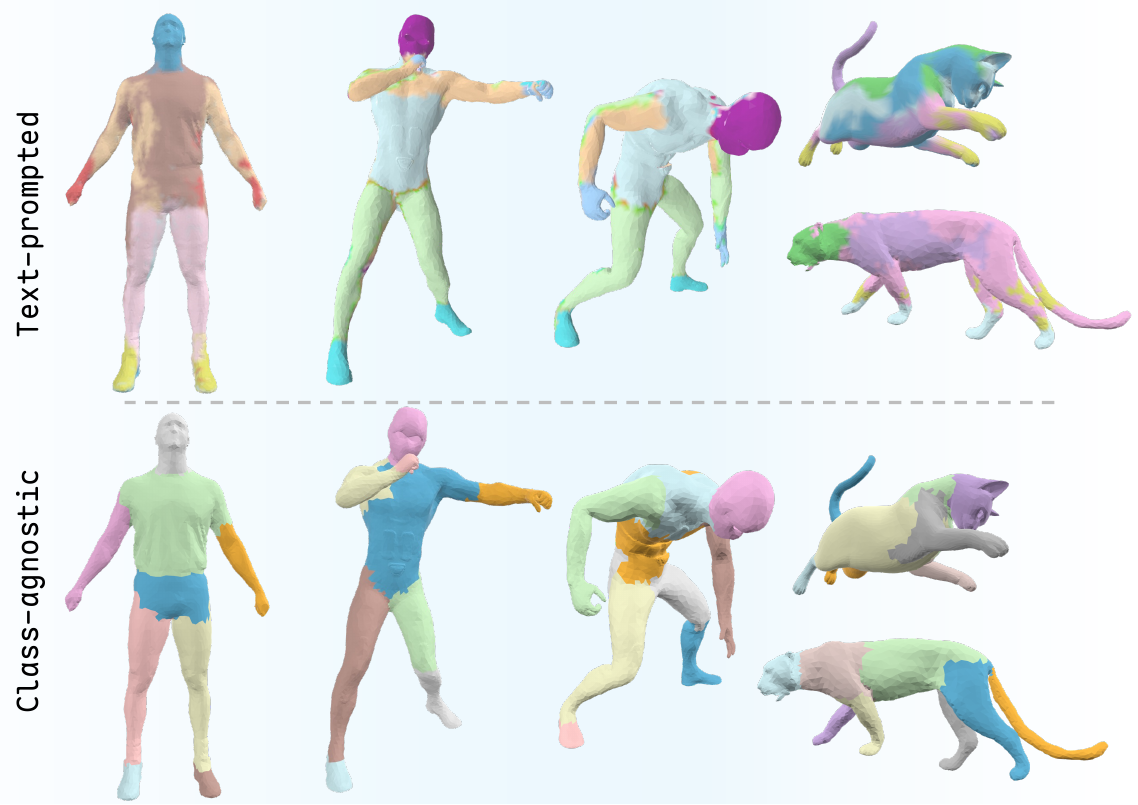}
    \caption{Comparison between class-agnostic segmentation PartField~\citep{liu2025partfield} and text-prompted segmentation SATR~\citep{abdelreheem2023satr}. Different parts are rendered in random colors.}
    \label{fig:seg_compare}
\end{figure}

\paragraph{Class-agnostic segmentation.} We consider an effective class-agnostic 3D segmentation method, PartField~\citep{liu2025partfield} to segment the 3D shape into non-intersecting parts. The number of semantic parts $n_\mathcal{R}$ is selected empirically to provide enough semantic information and avoid over-segmentation. We use $n_\mathcal{R}=9$ for human data and $n_\mathcal{R}=8$ for animal data. We also illustrate the segmentation results from PartField~\citep{liu2025partfield} and text-prompted method SATR~\citep{abdelreheem2023satr} in \cref{fig:seg_compare}. Compared with PartField, the segmentation results of SATR contain significant noise and ambiguous boundaries.

\paragraph{Semantic feature fields.} We first perform view-consistent texture synthesis, SyncMVD~\citep{liu2024text}, for uncolored shapes and render them into $K$ multi-view RGB images using PyTorch3D~\citep{ravi2020pytorch3d} ($K = 10$ for all experiments). We use uniformly distributed elevation and azimuthal angles in $[0^\circ,360^\circ)$. Then we extract semantic features using SD-DINO and employ the FeatUp upscaler~\citep{fu2024featup} to upsample them. Lastly, we back-project 2D semantic features into the 3D domain from all visible views using known camera parameters and average them to obtain per-vertex features.

\paragraph{Network and functional map.} We use DiffusionNet~\citep{sharp2022diffusionnet} as our feature refiner $\mathcal F$. The dimensions of $\bff_\text{geo}$ and $\bff_\text{sem}$ are $128$ and $768$ respectively. The output dimension of $\bff_\text{out}$ is $256$. For functional map calculation, we empirically set $\lambda_\text{couple}=1.0$ and $\lambda_\text{reg}=1.0$ in \cref{eq:fmaps}, following URSSM~\citep{cao2023unsupervised}.

\paragraph{Training.} We use the AdamW~\citep{loshchilov2017decoupled} optimizer with learning rate $10^{-3}$ for all experiments. We train our framework for $15$ epochs.

\section{Details of Datasets}
\label{suppl:dataset}

\subsection{Inter-Class Datasets}
\label{suppl:dataset_interclass}

\begin{figure}[ht]
    \centering
    \includegraphics[width=0.95\linewidth]{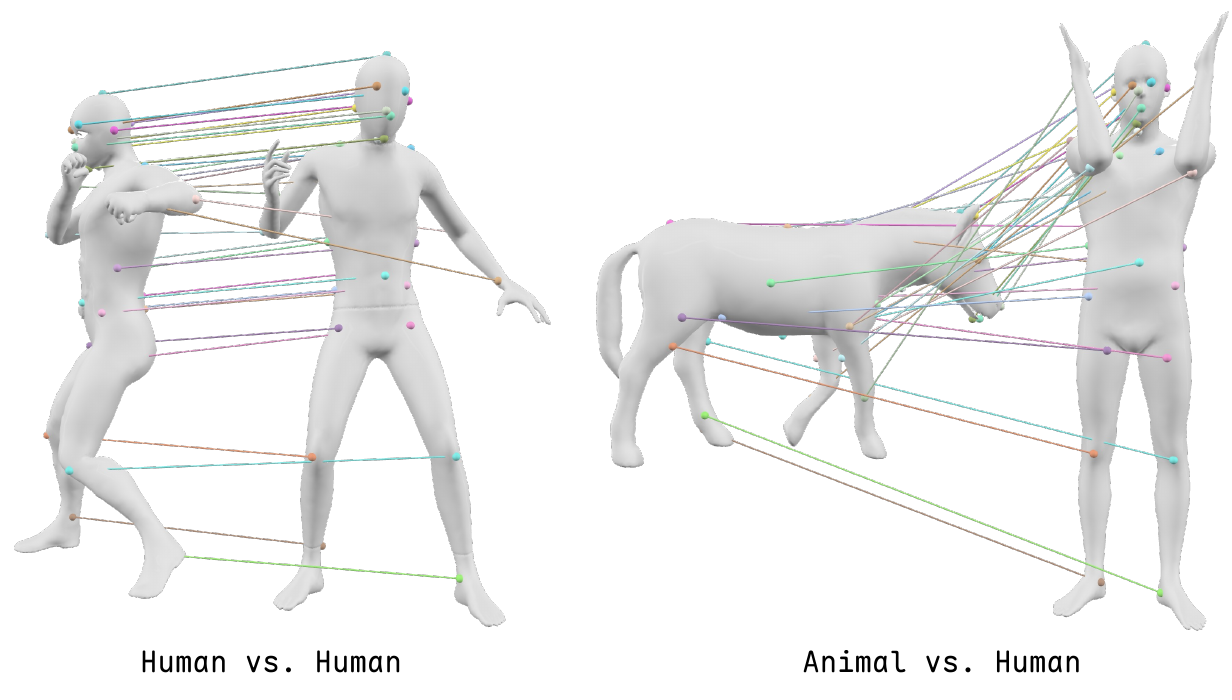}
    \caption{Example shape pairs and annotations from SNIS~\citep{abdelreheem2023zero}. SNIS contains strongly non-isometric and cross-category shape pairs, such as human vs. animal and human vs. humanoid, with semantically annotated keypoint correspondences.}
    \label{fig:snis}
\end{figure}

\begin{itemize}
    \item \textbf{SNIS}~\citep{abdelreheem2023zero} is a dataset to test strongly non-isometric and cross-category matching (\eg a human \vs a dog), containing 211 shapes from different sources, including FAUST \citep{bogo2014faust} (human), SMAL \citep{zuffi20173d} (animals), and DeformingThings4D (DT4D) \citep{li20214dcomplete} (humanoid). Annotations include 34 semantically corresponding keypoints for each shape. We use 250 test pairs provided by SNIS for evaluation (all test pairs are cross-category) and other pair combinations for training. \cref{fig:snis} shows some example pairs from SNIS.
    \item \textbf{TOSCA}~\citep{bronstein2008numerical} consists of synthetic high-resolution meshes of humans and animals captured in different poses. It contains 80 objects that span several categories, including cats, dogs, horses, centaurs, gorillas, humans, \etc. We remesh the original meshes to keep about $10,000$ faces. Since no cross-category annotations are provided, we prompted GPT-5 to find semantically matched keypoints between categories and finally filtered 20 keypoints. Similar with SNIS, we randomly select half samples from each category, and build the test pairs cross categories, result in $380$ pairs for evaluation. We use the rest pair combinations for training.
    \item \textbf{SHREC07}~\citep{Giorgi2007SHapeRC} consists of 400 watertight shapes and 20 semantic categories with 20 shapes per class, representing a wide range of objects such as humanoids, mechanical parts, animals, and furniture. Since some categories contain extremely sparse keypoints and share almost no common structures, \eg table, vase, and fish, we only conduct experiments on filtered categories, including human, teddy, armadillo, and four-legged animals, and prompted GPT-5 to find semantically matched keypoints between them, finally obtaining 15 keypoints. Similar to TOSCA, we randomly select half samples from each category, and build $100$ pairs for evaluation and the rest pair combinations for training.
\end{itemize}

\subsection{Non-Isometric Datasets}
\label{suppl:dataset_noniso}

\begin{itemize}
    \item \textbf{SMAL}~\citep{zuffi20173d} contains 49 four-legged animal shapes representing 8 species, such as dogs, cats, lions, horses, and \etc. Each shape is represented as a watertight mesh with about 8,000 faces. Following~\citet{cao2023unsupervised}, we use five species for training and three unseen species for evaluation.
    \item \textbf{TOPKIDS}~\citep{lahner2016shrec} is designed for evaluating non-isometric deformations and topological noises. It contains 26 shapes of synthetic human (``fat kid'') in various poses. Each shape includes ground-truth dense correspondences to a reference template shape.
\end{itemize}

\subsection{Near-Isometric Datasets}
\label{suppl:dataset_iso}

\begin{itemize}
    \item \textbf{FAUST}~\citep{bogo2014faust} consists of 10 human subjects with 10 different poses, where 80 shapes are provided for training and 20 shapes for testing. Following~\citet{cao2023unsupervised}, we consider the remeshed versions from~\citet{donati2020deep}. Each shape contains around 5,000 vertices with ground-truth dense correspondences to a template shape.
    \item \textbf{SCAPE}~\citep{anguelov2005scape} contains 71 human shapes with various poses of the same person. The dataset is split into 51 training shapes and 20 testing shapes.
    \item \textbf{SHREC19}~\citep{melzi2019shrec} is a benchmark to evaluate non-rigid shape matching, with a particular focus on different mesh connectivities in human shapes. It contains 44 human shapes with various identities and poses.
\end{itemize}

\section{Details of Baselines}
\label{suppl:baseline}

For URSSM~\citep{cao2023unsupervised}, we use the same training scheme and parameter settings with \ourmethod. \textit{To keep a fair comparison, we disable the test-time adaptation.} For Diff3F~\citep{dutt2024diffusion}, we follow the original implementation. Since the code of ZSC~\citep{abdelreheem2023zero} is not open-source, we implement it according to the original paper. We use Grounding-DINO~\citep{liu2024grounding} for visual grounding and SAM~\citep{kirillov2023segment} to obtain masks. Given segmentation results, the final dense correspondences are calculated by BCICP~\citep{ren2018continuous}. For DenseMatcher~\citep{zhu2025densematcher}, we use the same training scheme, parameter settings, and the same semantic features on textured meshes with \ourmethod. Since DenseMatcher requires predefined semantic parts, we use the same segmentation results of ZSC.

\section{Limitation and Future Work}

While \ourmethod demonstrates strong performance in universal 3D shape matching under non-isometric and inter-class scenarios, several limitations remain. First, GPT-5 may fail to correctly order symmetric or repetitive parts (\eg, chair legs), as semantic part names alone cannot infer geometric arrangement. Incorporating explicit object orientation cues or relational priors may alleviate such issues. Second, our method still needs a separate procedure to extract semantic features using vision foundation models. Training a unified feed-forward feature extractor that distills visual knowledge from foundation models, rather than a feature refiner, may address this issue. Finally, the current framework relies on multimodal large language models for prompting during training-time part naming to enable semantic awareness. Extending this to more efficient pipelines could improve scalability and practicality. Future work will address these challenges by exploring enhanced structural priors of shapes, a unified feed-forward feature extractor, and reducing model intervention during matching.

\end{document}